\documentclass[10pt, a4paper]{article}
\usepackage{lrec}
\usepackage{multibib}
\newcites{languageresource}{Language Resources}
\usepackage{graphicx}
\usepackage{tabularx}
\usepackage{soul}

\usepackage{epstopdf}
\usepackage[utf8]{inputenc}

\usepackage{hyperref}
\usepackage{xstring}

\usepackage{color}

\usepackage{multirow}
\usepackage{mathtools}
\usepackage{times}
\usepackage{latexsym}

\usepackage{verbatim}

\usepackage{todonotes}

\title{\textbf{Adaptation of a Lexical Organization for\\ Social Engineering Detection and Response Generation}}

\name{\begin{tabular}{c}Archna Bhatia, Adam Dalton, Brodie Mather, Sashank Santhanam, \\
Samira Shaikh, Alan Zemel, Tomek Strzalkowski, Bonnie J. Dorr\end{tabular}}

\address{The Florida Institute for Human and Machine Cognition, The University of North Carolina at Charlotte, \\University of Albany NY, Rensselaer Polytechnic Institute NY \\
         \{abhatia,adalton,bmather,bdorr\}@ihmc.us, \{ssantha1,sshaikh2\}@uncc.edu, \\azemel@albany.edu,
         tomek@rpi.edu\\}

\abstract{
We present a paradigm for extensible lexicon development based on Lexical Conceptual Structure to support social engineering detection and response generation. We leverage the central notions of \textit{ask} (elicitation of behaviors such as providing access to money) and \textit{framing} (risk/reward implied by the ask). We demonstrate improvements in ask/framing detection through refinements to our lexical organization and show that response generation qualitatively improves as ask/framing detection performance improves. The paradigm presents a systematic and efficient approach to resource adaptation for improved task-specific performance.\\ \newline  }

\begin{document}
\maketitleabstract

\section{Introduction}

Social engineering (SE) refers to sophisticated use of deception to manipulate individuals into divulging confidential or personal information for fraudulent purposes. Standard cybersecurity defenses are ineffective because attackers attempt to exploit humans rather than system vulnerabilities. Accordingly, we have built a \textit{user alter-ego} application that detects and engages a potential attacker in ways that expose their identity and intentions. 

Our system relies on a paradigm for extensible lexicon development that leverages the central notion of \textit{ask}, i.e., elicitation of behaviors such as PERFORM (e.g., clicking a link) or GIVE (e.g., providing access to money).  This paradigm also enables detection of risk/reward (or LOSE/GAIN) implied by an ask, which we call \textit{framing} (e.g., \textit{lose your job}, \textit{get a raise}). These elements are used for countering attacks through bot-produced responses and actions. The system is tested in an email environment, but is applicable to other forms of online communications, e.g., SMS. 

\begin{table}[!h]
\begin{center}
    \begin{small}
    \begin{tabular}{|p{1.6in}|p{.5in}|p{.4in}|}\hline
     \textbf{ Email}  & \textbf{Ask} & \textbf{Framing} \\ \hline

    (a) It is a pleasure to inform you that you have won 1.7Eu. Contact me. ({\color{blue}\underline{jw11@example.com}})
      & PERFORM contact \newline ({\color{blue}\underline{jw11@...}}) 
      & GAIN \newline won \newline (1.7Eu)
      
     \\ \hline

    (b) You won \$1K. Did you send money? Do that by 9pm or lose money. Respond asap.
      & GIVE \newline send \newline(money)
      & LOSE \newline lose \newline (money) 
      
    \\ \hline
    
    (c) Get 20\% discount. Check {\color{blue} \underline{eligibility}} or paste this link: {\color{blue}\underline{http...}}. Sign up for \color{blue}\underline{email alerts}.
      & PERFORM \newline paste \newline({\color{blue}\underline{http...}})
      & GAIN \newline get \newline (20\%) 
      
    \\ \hline

    \end{tabular}
    \end{small}
    \vspace*{-.1in}
    \caption{LCS+ Ask/Framing output for three SE emails}
    \label{tab:Examples}
    \vspace*{-.25in}
\end{center}
\end{table} 

More formally, an \textit{ask} is a statement that elicits a behavior from a potential victim, e.g., \textit{please buy me a gift card}. Although asks are not always explicitly stated \cite{Drew2014Requesting,Zemel2017Texts}, we discern these through navigation of semantically classified verbs. The task of ask detection specifically is targeted event detection based on parsing and/or Semantic Role Labeling (SRL), to identify semantic class triggers \cite{dorr:2020}. \textit{Framing} sets the stage for the ask, i.e., the purported threat (LOSE) or benefit (GAIN) that the social engineer wants the potential victim to believe will obtain through compliance or lack thereof. It should be noted that there is no one-to-one ratio between ask and framing in the ask/framing detection output. Given the content, there may be none, one or more asks and/or framings in the output.

Our lexical organization is based on \textit{Lexical Conceptual Structure} (LCS), a formalism that supports resource construction and extensions to new applications such as SE detection and response generation. Semantic classes of verbs with similar meanings (\textit{give}, \textit{donate}) are readily augmented through adoption of the STYLUS variant of LCS \citelanguageresource{dorr-voss:2018} and \cite{Dor:18a}. 
We derive LCS+ from asks/framings and employ CATVAR \citelanguageresource{Nizar:2003} to relate word variants (e.g., \textit{reference} and \textit{refer}). 
Table~\ref{tab:Examples} illustrates LCS+ Ask/Framing output for three (presumed) SE emails: two PERFORM asks and one GIVE ask.\footnote{To view our system's ask/framing outputs on a larger dataset (the same set of emails which were also used for ground truth (GT) creation described below), refer to \url{https://social-threats.github.io/panacea-ask-detection/data/case7LCS+AskDetectionOutput.txt}.} Parentheses () refer to ask \textit{arguments}, often a link that the potential victim might choose to click.  Ask/framing outputs are provided to downstream response generation. For example, a possible response for Table~\ref{tab:Examples}(a) is \textit{I will contact asap.}

A comparison of LCS+ to two related resources shows that our lexical organization supports refinements, improves ask/framing detection and top ask identification, and yields qualitative improvements in response generation. LCS+ is deployed in a SE detection and response generation system. Even though LCS+ is designed for the SE domain, the approach to development of LCS+ described in this paper serves as a guideline for developing similar lexica for other domains. Correspondingly, even though development of LCS+ is one of the contributions of this paper, the main contribution is not this resource but the systematic and efficient approach to resource adaptation for improved task-specific performance.

\section{Method}
\label{sec:method}

In our experiments described in Section \ref{sec:results}, we compare LCS+, our lexical resource we developed for the SE domain, against two strong baselines: STYLUS and Thesaurus.\\


\textbf{STYLUS baseline:} 
As one of the baselines for our experiments, we leverage a publicly available resource STYLUS that is based on Lexical Conceptual Structure (LCS) \citelanguageresource{dorr-voss:2018} and \cite{Dor:18a}. The LCS representation is an underlying representation of spatial and motion predicates \cite{Jackendoff:1983,Jackendoff:1990,Dorr:1993}, such as \textit{fill} and \textit{go}, and their metaphorical extensions, e.g., temporal (the hour \textit{flew} by) and possessional (he \textit{sold} the book).\footnote{LCS is publicly available at \url{https://github.com/ihmc/LCS}.} 
Prior work \cite{Jackendoff:1996,Levin:1993,Olsen:1994,Chang-Patent:LOM2007,Chang-Patent:LSS2010,Kipper:2007,Palmer:2017} has suggested that there is a close relation between underlying lexical-semantic structures of verbs and nominal predicates and their syntactic argument structure. We leverage this relationship to extend the existing STYLUS verb classes for the resource adaptation to SE domain through creation of LCS+ which is discussed below.  

For our STYLUS verb list, we group verbs into four lists based on asks (PERFORM, GIVE) and framings (LOSE, GAIN). The STYLUS verb list can be accessed here: \url{https://social-threats.github.io/panacea-ask-detection/resources/original_lcs_classes_based_verbsList.txt}. Examples of this classification
 are shown below (with total verb count in parentheses):

\begin{itemize}
    \item{PERFORM (214): remove, redeem, refer}
    \item{GIVE (81): administer, contribute, donate}
    \item{LOSE (615): penalize, stick, punish, ruin}
    \item{GAIN (49): accept, earn, grab, win}
\end{itemize}

Assignment of verbs to these four ask/framing categories is determined by a computational linguist, with approximately a person-day of human effort. Identification of genre-specific verbs is achieved through analysis of 46 emails (406 clauses) after parsing/POS/SRL is applied. 

As an example, the verb \textit{position} (Class 9.1) and the verb \textit{delete} (Class 10.1) both have an underlying \textit{placement} or \textit{existence} component with an affected object (e.g., the cursor in \textit{position your cursor} or the account in \textit{delete your account}), coupled with a location (e.g., \textit{here} or \textit{from the system}).  Accordingly, \textit{Put} verbs in Class 9.1 and \textit{Remove} verbs in Class 10.1 are grouped together and aligned with a PERFORM ask (as are many other classes with similar properties: Banish, Steal, Cheat, Bring, Obtain, etc.). Analogously, verbs in the \textit{Send} and \textit{Give} classes are aligned with a GIVE ask, as all verbs in these two classes have a sender/giver and a recipient. 

Lexical assignment of framings is handled similarly, i.e., verbs are aligned with LOSE and GAIN according to their argument structures and components of meaning. It is assumed that the potential victim of a SE attack serves to lose or gain something, depending on non-compliance or compliance with a social engineer's ask.  As an example, the framing associated with the verb \textit{losing} (Class 10.5) in \textit{Read carefully to avoid \textbf{losing} account access} indicates the risk of losing access to a service; Class 10.5 is thus aligned with LOSE. Analogously, the verb \textit{win}  (Class 13.5.1) in \textit{You have \textbf{won} 1.7M Eu.} is an alluring statement with a purported gain to the potential victim; thus Class 13.5.1 is aligned with GAIN.  In short, verbs in classes associated with LOSE imply negative consequences (Steal, Impact by Contact, Destroy, Leave) whereas verbs in classes associated with GAIN imply positive consequences (Get, Obtain).  

Some classes are associated with more than one ask/framing category: Steal (Class 10.5) and Cheat (Class 10.6) are aligned with both PERFORM (\textit{redeem, free}) and LOSE (\textit{forfeit, deplete}). Such distinctions are not captured in the lexical resource, but are algorithmically resolved during ask/framing detection, where contextual clues provide disambiguation capability. For example, \textit{Redeem coupon} is a directive with an implicit request to click a link, i.e., a PERFORM. By contrast, \textit{Avoid losing account access} is a statement of risk, i.e., a LOSE. The focus here is not on the processes necessary for distinguishing between these contextually-determined senses, but on the organizing principles underlying both
, in support of application-oriented resource construction.\\

\textbf{LCS+ resource for SE adapted from STYLUS:} Setting disambiguation aside, resource improvements are still necessary for the SE domain because, due to its size and coverage, STYLUS is likely to predict a large number of both true and false positives during ask/framing detection. To reduce false positives without taking a hit to true positives, we leverage an important property of the LCS paradigm: its extensible organizational structure wherein similar verbs are grouped together. With just one person-day of effort by two computational linguists (authors on the paper; the algorithm developer, also an author, was not involved in this process), a new lexical organization, referred to as ``LCS+'' is derived from STYLUS, taken together with asks/framings from 
a set of 46 malicious/legitimate emails.\footnote{It should be noted that this resource adaptation is based on an analysis of emails not related to, and without access to, the adjudicated ground truth described in section \ref{sec:results} That is, the 46 emails used for resource adaptation are distinct from the 20 emails used for creating adjudicated ground truth.} These emails are a random subset of 1000+ emails (69 malicious and 938 legitimate) sent from an external red team to five volunteers in a large government agency using social engineering tactics. Verbs from these emails are tied into particular LCS classes with matching semantic peers and argument structures. These emails are proprietary but the resulting lexicon is released here: \url{https://social-threats.github.io/panacea-ask-detection/resources/lcsPlus_classes_based_verbsList.txt}. 

Two categories (PERFORM and LOSE) are modified from the adaptation of LCS+ beyond those in STYLUS:
\begin{itemize}
    \item{PERFORM (6 del, 44 added): copy, notify}
    \item{GIVE (no changes)}
    \item{LOSE (174 del, 11 added): forget, surrender}
    \item{GAIN (no changes)}
\end{itemize}

Table~\ref{tab:full-set-of-modifications} shows the refined lexical organization for LCS+ with ask categories (PERFORM, GIVE) and framing categories (GAIN, LOSE). Boldfaced class numbers indicate the STYLUS classes that were modified. The resulting LCS+ resource drives our SE detection/response system. Each class includes italicized examples with boldfaced triggers. The table details changes to PERFORM and LOSE categories. For PERFORM, there are 6 deleted verbs across 10.2 (Banish Verbs) and 30.2 (Sight Verbs) and also 44 new verbs added to 30.2.  For LOSE, 7 classes are associated with additions and/or deletions, as detailed in the table.\\

\textbf{Thesaurus baseline:} The Thesaurus baseline is based on an expansion of simple forms of framings. Specifically, the verbs \textit{gain}, \textit{lose}, \textit{give}, and \textit{perform}, are used as search terms to find related verbs in a standard but robust resource \url{thesaurus.com} (referred to as ``Thesaurus''). The verbs thus found are grouped into these same four categories: 


\begin{itemize}
    \item{PERFORM (44): act, do, execute, perform}
    \item{GIVE (55): commit, donate, grant, provide}
    \item{LOSE (41): expend, forefeit, expend, squander}
    \item{GAIN (53): clean, get, obtain, profit, reap}
\end{itemize}

The resulting Thesaurus verb list is publicly released here: \url{https://social-threats.github.io/panacea-ask-detection/resources/thesaurus_based_verbsList.txt}.\\

We also adopt categorial variations through  CATVAR  \citelanguageresource{Nizar:2003} to map between different parts of speech, e.g., \textit{winner}(N) $\rightarrow$ \textit{win}(V). STYLUS, LCS+ and Thesaurus  contain verbs only, but asks/framings are often nominalized. For example, \textit{you can reference your gift card} is an implicit ask to examine a gift card, yet without CATVAR this ask is potentially missed. CATVAR recognizes \textit{reference} as a nominal form of \textit{refer}, thus enabling the identification of this ask as a PERFORM.


\section{Experiments and Results} \label{sec:results}

\textit{Intrinsic} evaluation of our resources is based on comparison of ask/framing detection to an adjudicated ground truth (henceforth, GT), a set of 472 clauses from system output on 20 unseen emails. These 20 emails are a random subset of 2600+ messages collected in an email account set up to receive messages from an internal red team as well as ``legitimate'' messages from corporate and academic mailing lists. As alluded to earlier, these 20 emails are distinct from the dataset used for resource adaptation to produce the task-related LCS+. 

The GT is produced through human adjudication and correction by a computational linguist\footnote{The adjudicator is an author but is not the algorithm developer, who is also an author.} of initial ask/framing labels automatically assigned by our system to the 472 clauses. System output also includes the identification of a ``top ask'' for each email, based on the degree to which ask argument positions are filled.\footnote{Argument positions express information such as the ask type (i.e. PERFORM), context to the ask (i.e. financial), and the ask target (e.g., ``you'' in ``Did you send me the money?'').}
\textit{Top asks} are adjudicated by the computational linguist once the ask/framing labels are adjudicated. The resulting GT is accessible here: \url{https://social-threats.github.io/panacea-ask-detection/data/}. 

The GT is used to measure the precision/recall/F of three of three variants of ask detection output (Ask, Framing, and Top Ask) corresponding to our three lexica: Thesaurus, STYLUS, and LCS+. LCS+ is favored (with statistical significance) against the two very strong baselines, Thesaurus and STYLUS.  Table~\ref{Tab:Results1} presents results: Recall for framings is highest for STYLUS, but at the cost of higher false positives (lower precision). F-scores increase for STYLUS over Thesaurus, and for LCS+ over STYLUS. 

McNemar \cite{McNemar47samplingIndependence} tests yield statistically significant differences for asks/framings at the 2\% level between Thesaurus and LCS+ and between STYLUS and LCS+.\footnote{Tested values were TP$+$TN vs FP$+$FN, i.e., significance of change in total error rate.} It should be noted that not all clauses in GT are ask or framing: vast majority (80\%) are neither (i.e., they are true negatives). 

We note that an alternative to the Thesaurus and LCS baselines would be a bag-of-words lexicon, with no organizational structure.  However, the key contribution of this work is the ease of adaptation through classes, obviating the need for training data (which are exceedingly difficult to obtain). Classes enable extension of a small set of verbs to a larger range of options, e.g., if the human determines from a small set of task-related emails that \textit{provide} is relevant, the task-adapted lexicon will include \textit{administer}, \textit{contribute}, and \textit{donate} for free. If a class-based lexical organization is replaced by bag-of-words, we stand to lose efficient (1-person-day) resource adaptation and, moreover, training data would be needed.  

\begin{table}
\begin{center}
    \begin{small}
    \begin{tabular}{|p{2.7in}|}
\multicolumn{1}{c}{\textbf{PERFORM:}}\\
\hline
9.1 Put Verbs: \textit{\textbf{Position} your cursor \underline{here}}\\
10.1 Remove Verbs: \textit{\textbf{Delete} virus from machine}\\
\textbf{10.2} Banish Verbs$\rightarrow$5 deleted (banish, deport, evacuate, extradite, recall): \textit{\textbf{Remove} fee from your account}\\
10.5 Steal Verbs: \textit{\textbf{Redeem} coupon below}\\
10.6 Cheat Verbs: \textit{\textbf{Free} yourself from debt}\\
11.3 Bring and Take Verbs: \textit{\textbf{Bring} me a gift card}\\
13.5.2 Obtain: \textit{\textbf{Purchase} two gift cards}\\
\textbf{30.2} Sight Verbs$\rightarrow$1 deleted (regard), 44 added (e.g., check, eye, try, view, visit): \textit{\textbf{View} this \underline{website}}\\
37.1 Transfer of Message: \textit{\textbf{Ask} for a refund}\\
37.2 Tell Verbs: \textit{\textbf{Tell} them \$50 per card}\\
37.4 Communication: \textit{\textbf{Sign} the back of the card}\\
42.1 Murder Verbs: \textit{\textbf{Eliminate} your debt \underline{here}}\\
44 Destroy Verbs: \textit{\textbf{Destroy} the card}\\
54.4 Price Verbs: \textit{\textbf{Calculate} an amount \underline{here}}\\ 
\hline
\multicolumn{1}{c}{\textbf{~}}\\
\multicolumn{1}{c}{\textbf{GIVE:}}\\
\hline
11.1 Send Verbs: \textit{\textbf{Send} me the gift cards}\\
13.1 Give Verbs: \textit{\textbf{Give} today}\\
13.2 Contribute Verbs: \textit{\textbf{Donate}!}\\
13.3 Future Having: \textit{\textbf{Advance} me \$100}\\
13.4.1 Verbs of Fulfilling: \textit{\textbf{Credit} your account}\\
32.1 Want Verbs: \textit{I \textbf{need} three gift cards}\\ 
\hline
\multicolumn{1}{c}{\textbf{~}}\\
\multicolumn{1}{c}{\textbf{LOSE:}}\\
\hline
\textbf{10.5} Steal Verbs$\rightarrow$11 added (e.g., forfeit, lose, relinquish, sacrifice): \textit{Don't \textbf{forfeit} this chance!}\\
10.6 Cheat Verbs: \textit{Are your funds \textbf{depleted}?}\\
17.1 Throw Verbs: \textit{Don't \textbf{toss} out this coupon}\\
17.2 Pelt Verbs: \textit{Scams \textbf{bombarding} you?}\\
18.1 Hit Verbs: \textit{Don't be \textbf{beaten} by debt}\\
18.2 Swat Verbs: \textit{\textbf{Sluggish} market getting you down?}\\
18.3 Spank Verbs: \textit{\textbf{Clobbered} by fees?}\\
18.4 Impact by Contact: \textit{Avoid being \textbf{hit} by malware}\\
19 Poke Verbs: \textit{\textbf{Stuck} with debt?}\\
\textbf{29.2} Characterize Verbs$\rightarrow$16 deleted (e.g., appreciate, envisage): \textit{\textbf{Repudiated} by creditors?}\\
\textbf{29.7} Orphan Verbs$\rightarrow$5 deleted (apprentice, canonize, cuckold, knight, recruit): \textit{Avoid \textbf{crippling} debt}\\
\textbf{29.8} Captain Verbs$\rightarrow$35 deleted (e.g., captain, coach, cox, escort): \textit{\textbf{Bullied} by bill collectors?}\\
\textbf{31.1} Amuse Verbs$\rightarrow$91 deleted (e.g., amaze, amuse, gladden): \textit{Don't be \textbf{disarmed} by hackers}\\
\textbf{31.2} Admire Verbs$\rightarrow$26 deleted (e.g., admire, exalt); \textit{Are you \textbf{lamenting} your credit score?}\\
\textbf{31.3} Marvel Verbs$\rightarrow$1 deleted (feel): \textit{Living in \textbf{fear}?}\\
33 Judgment Verbs: \textit{Need to remove \textbf{penalties}?}\\
37.8 Complain Verbs: \textit{Want your \textbf{gripes} answered?}\\
42.1 Murder Verbs: \textit{Debt \textbf{killing} your credit?}\\
42.2 Poison Verbs: \textit{\textbf{Strangled} by debt?}\\
44 Destroy Verbs: \textit{PC \textbf{destroyed} by malware?}\\
48.2 Disappearance: \textit{Your account will \textbf{expire}}\\
51.2 Leave Verbs: \textit{Found your \textbf{abandoned} prize}\\
\hline 
\multicolumn{1}{c}{~}\\ 
\multicolumn{1}{c}{\textbf{GAIN:}}\\ 
\hline
13.5.1 Get: \textit{You are a \textbf{winner} of 1M Eu.}\\
13.5.2 Obtain: \textit{You can \textbf{recover} your credit rating}\\ \hline
    \end{tabular}
    \end{small}
    \caption{Lexical organization of LCS+ relies on Ask Categories (PERFORM, GIVE) and Framing Categories (GIVE, LOSE). Italicized exemplars with boldfaced triggers illustrate usage for each class. Boldfaced class numbers indicate those STYLUS classes that were modified to yield the LCS+ resource.}
    \label{tab:full-set-of-modifications}
\end{center}
\end{table}

\begin{table}[!h]
\begin{center}
\begin{small}
\begin{center}
\begin{tabular}{|l|r|r|r|}\hline
\textbf{Thesaurus}&\textbf{P}&\textbf{R}&\textbf{F}\\ \hline
Ask:    &0.273 & 0.042 & 0.072 \\
Framing:&0.265 & 0.360 & 0.305  \\
TopAsk: &0.273 & 0.057 & 0.094  \\
\hline
\hline
\textbf{STYLUS}&\textbf{P}&\textbf{R}&\textbf{F}\\ \hline
Ask:&  0.333 & 0.104 & 0.159  \\
Framing:& 0.298 & \textbf{0.636} & 0.406  \\
TopAsk:&  0.571 & 0.151 & 0.239  \\
\hline
\hline
\textbf{LCS+}&\textbf{P}&\textbf{R}&\textbf{F}\\ \hline
Ask:&       \textbf{0.667} & \textbf{0.411} & \textbf{0.508}  \\
Framing: &  \textbf{0.600} & 0.600 & \textbf{0.600}  \\
TopAsk:&    \textbf{0.692} & \textbf{0.340} & \textbf{0.456}  \\
\hline
\end{tabular}
\end{center}
\end{small}
\vspace*{-.05in}
\caption{\label{Tab:Results1} Impact of lexical resources on ask/framing detection: Thesaurus, STYLUS, LCS+
}
\vspace*{-.13in}
\end{center}
\end{table}

A first step toward \textit{extrinsic} evaluation is inspection of responses generated from each resource's top ask/framing pairs. Table~\ref{tab:Examples} (given earlier) shows LCS+ ask/framing pairs whose corresponding (T)hesaurus and (S)TYLUS pairs are:\\

\begin{footnotesize}
(a) 
\textbf{T}: None, None \\
\mbox{~~~~~~}\textbf{S}: None, GAIN/won(1.7Eu)\\
(b) 
\textbf{T}: PERFORM/do(that), LOSE/lose(money)\\
\mbox{~~~~~~}\textbf{S}: GAIN/won(money), GIVE/send(money)\\
(c) \textbf{T}: None, GAIN/get(20\%)\\
\mbox{~~~~~~}\textbf{S}: PERFORM/sign(http:..), GAIN/get(20\%)\\
\end{footnotesize}

Below are corresponding examples of generated responses\footnote{For brevity, \textit{excerpts} are shown in lieu of full emails.} for all 3 resources, based on a templatic approach that leverages ask/framing hierarchical structure and corresponding confidence scores. This module is part of a larger, separate publication.\\

\begin{footnotesize}
(a) 
\textbf{T}: How are you? Thanks. \\
\mbox{~~~~~~}\textbf{S}: ...too good to be true. What should I do?\\
\mbox{~~~~~~}\textbf{L+}: I will contact asap. \\
(b) 
\textbf{T}: Thanks for getting in touch, need more info.\\
\mbox{~~~~~~}\textbf{S}: Nervous about this. Your name?\\
\mbox{~~~~~~}\textbf{L+}: I would respond,\footnote{LCS+ detects both GIVE/send and PERFORM/respond.} but I need more info.\\
(c) 
\textbf{T}: What should I do now?\\
\mbox{~~~~~~}\textbf{S}: Website doesn't open, is this the link?\\
\mbox{~~~~~~}\textbf{L+}: Thanks, need more info before I paste link\\
\end{footnotesize}

There are qualitative differences in these responses. For example, in (a) Thesaurus (T) yields no asks/framings; thus a canned response is generated. By contrast, the same email yields a more responsive output for STYLUS (S), and a more focused response for LCS+ (L). Similar distinctions are found for responses in (b) and (c). Note that in the LCS+ condition, if there is no match found using LCS+, downstream response generation prompts the attacker (e.g., ``please clarify'') until an interpretable ask or framing appears. In this SE task, not all responses move the conversation forward. A central goal of the SE task is to waste the attacker's time, play along, and possibly extract information that could unveil their identity.

\section{Related Work}
\label{sec:rel}

LCS is used in interlingual machine translation \cite{Voss:1995,Habash:2002}, lexical acquisition \cite{Habash:2006}, cross-language information retrieval \cite{Levow:2000}, language generation \cite{Traum:2000}, and intelligent language tutoring \cite{Dorr-icall:1997}. STYLUS \citelanguageresource{dorr-voss:2018} and \cite{Dor:18a} systematizes LCS based on several studies \cite{Levin:1995,Levin:1998}, but to our knowledge our work is the first use of LCS in a conversational context, within a cyber domain.

Our approach relates to work on conversational agents (CAs), where neural models automatically generate responses \cite{gao2019neural,santhanam2019survey}, topic models produce focused responses \cite{dziri2018augmenting},
self-disclosure yields targeted responses \cite{RavichanderB18}, and SE detection employs topic models \cite{Bhakta:2015} and NLP of conversations \cite{Yuki:2016}. However, all such approaches are limited to a pre-defined set of topics, constrained by the training corpus. 

Other prior work focuses on persuasion detection/ prediction \cite{mckeown:aaai-2018} by leveraging argument structure, but for the purpose of judging when a persuasive attempt might be successful in subreddit discussions dedicated to changing opinions (ChangeMyView). Our work aims to achieve effective dialogue for countering (rather than adopting) persuasive attempts. 

Text-based semantic analysis for SE detection \cite{kimcatch} is related to our work but differs in that our work focuses not just on \textit{detecting} an attack, but on \textit{engaging} with an attacker.
Whereas a bot might be employed to warn a potential victim that an attack is underway, our bots are designed to communicate with a social engineer in ways that elicit identifying information. 

\section{Conclusions} \label{sec:conclusions}

Both STYLUS and LCS+ support ask/framing detection in service of bot-produced responses. Intrinsically, LCS+ is superior to both STYLUS and Thesaurus when measured against human-adjudicated output, verified for significance by McNemar tests at the 2\% level. Extrinsically, STYLUS supports more responsive bot outputs and LCS+ supports more focused bot outputs. 

A more general advantage of adapting LCS+ to the SE domain is that it can act as a guideline for developing similar resources for other domains which will similarly support focused outputs appropriate for particular domains. The main contribution of this paper is not development of a particular task-specific resource, nor to suggest that LCS+ is a generic resource for many tasks, but to present a systematic, efficient approach to resource adaptation technique that can generalize to other tasks for improved task-specific performance, e.g., understanding viewpoints in social media or detecting motives behind activities of political groups. 
We acknowledge that our extrinsic evaluation is limited. While we have demonstrated the efficacy of  ask detection approaches on a set of representative emails, a quantitative evaluation is required to test the statistical significance of our extrinsic observations. Future work is planned to conduct experiments with crowd-sourced workers judging the efficacy and effectiveness of generated responses. 

\section*{Acknowledgments}
This work was supported by DARPA
through AFRL Contract FA8650-18-
C-7881 and through Army Contract
W31P4Q-17-C-0066. All statements of
fact, opinion or conclusions contained
herein are those of the authors and
should not be construed as representing
the official views or policies of DARPA,
AFRL, Army, or the U.S. Government.

\section{Bibliographical References}\label{reference}

\bibliographystyle{lrec}
\bibliography{lrec2020W,aaai2020}

\section{Language Resource References}
\label{lr:ref}
\bibliographystylelanguageresource{lrec}
\bibliographylanguageresource{languageresource}

\end{document}